\documentclass[sigconf]{acmart}

\usepackage{booktabs}
\usepackage{amsmath}
\usepackage{pifont}
\usepackage{enumitem}
\usepackage[most]{tcolorbox}
\usepackage{pbalance}

\definecolor{promptback}{HTML}{EFF4FC}
\definecolor{promptframe}{HTML}{A3C6F4}
\definecolor{prompttitle}{HTML}{D2E5FC}
\newtcolorbox{promptbox}[1]{enhanced, breakable, skin first=enhanced, skin middle=enhanced, skin last=enhanced,
  pad at break*=4pt, colback=promptback, colframe=promptframe, colbacktitle=prompttitle,
  coltitle=black, boxrule=0.6pt, arc=2pt, left=5pt, right=5pt, top=3pt, bottom=3pt, title={#1},
  fonttitle=\sffamily\bfseries\small, fontupper=\ttfamily\raggedright,
  before upper={\setlength{\parskip}{0.4\baselineskip}}}
\newcommand{\cond}[1]{{\normalfont\itshape\color{black!55}[#1]}}

\newcommand{\finding}[2]{\noindent\textbf{Finding #1: #2}}

\setcopyright{cc}
\setcctype[4.0]{by-nc}
\makeatletter
\let\orig@copyrightpermission\@copyrightpermission
\def\@copyrightpermission{$^{*}$Corresponding author: Yulong Bian.\par\orig@copyrightpermission}
\patchcmd{\maketitle}{{\itshape \acmConference@shortname, \acmConference@venue}\par}{}{}{\errmessage{venue patch failed}}
\makeatother
\acmConference{}{}{}
\acmBooktitle{}
\acmDOI{}
\acmISBN{}

\begin{document}
\pagestyle{plain}
\setlength{\emergencystretch}{1.5em}

\title{Words or Deeds? How LLMs Predict Users' Decisions in Human--Agent Dialogue}

\author{Baiqiao~Zhang$^{\spadesuit}$, Shuai~Ma$^{\diamondsuit}$, Zhifeng~Liao$^{\clubsuit}$, Chao~Zhou$^{\diamondsuit}$, Jianshu~Wang$^{\diamondsuit}$, Yuhan~Zhao$^{\clubsuit}$, Chaoyue~Zhang$^{\clubsuit}$, Tongjun~Si$^{\clubsuit}$, Yuying~Tang$^{\spadesuit}$, Xiangxian~Li$^{\clubsuit}$, Yong-Jin~Liu$^{\heartsuit}$, Yulong~Bian$^{\clubsuit *}$}
\affiliation{
  \institution{$^{\spadesuit}$\,The Hong Kong University of Science and Technology, Hong Kong SAR, China\\$^{\diamondsuit}$\,Institute of Software, Chinese Academy of Sciences, Beijing, China\\$^{\clubsuit}$\,Shandong University, Weihai, China\\$^{\heartsuit}$\,Tsinghua University, Beijing, China\\
  bzhangcw@cse.ust.hk, \{mashuai, zhouchao\}@iscas.ac.cn, yuying.tang@connect.ust.hk, jeffwang825@henu.edu.cn,\\
  \{202200810061, 202400800660, 202400800527, 202400800539\}@mail.sdu.edu.cn,\\
  \{xiangxianli, bianyulong\}@sdu.edu.cn, liuyongjin@tsinghua.edu.cn}
  \country{}
}
\renewcommand{\shortauthors}{Zhang et al.}

\begin{abstract}
Web systems increasingly predict users' decisions from what users say, yet when interests partly conflict, users may promise to cooperate to win the other side's cooperation and then not do so. Systems that rely on such promises, such as sales or service agents deciding what to offer next, may grant concessions to users who then do not keep their promises, at a cost to the party each agent represents. We test how LLMs combine users' statements with decision records to predict users' decisions, using 1,260 rounds in which 42 users played a repeated prisoner's dilemma-style game with five agents, each prompted to express a dominant Big Five trait, and could talk with the agent before every decision. Across eight open-weight LLMs, predictions from decision records separated cooperation from defection well (AUC = .80), and adding dialogue brought no detectable gain. With dialogue, predictions agreed with users' statements more often, and errors concentrated on rounds in which users broke their statements: accuracy on rounds with a statement barely changed (.73 to .74), while it rose on kept statements (.76 to .95) and fell sharply on broken ones (.65 to .17). Predictions weighted statements more heavily than users' behavior warranted: when users stated a decision they had not yet made in that game, the predicted probability of keeping it rose from .13 to .69, whereas users kept .40 of such statements. Users' decisions also varied with the agent they faced, but predictions hardly reflected this: users cooperated more with the Agreeableness agent than with the Extraversion agent (70.6\% vs.\ 54.5\%), whereas predicted cooperation was nearly identical (64.5\% vs.\ 63.1\%). Users also cooperated more with an agent when they themselves scored higher on that agent's trait (e.g., agreeable users with the Agreeableness agent), whereas predicted cooperation barely changed with these scores. Dialogue-based predictions should therefore weigh users' statements by their earlier decisions and the agent they face.
\end{abstract}



\begin{teaserfigure}
\centering
\includegraphics[width=\textwidth,trim={0 6.8bp 0 6.8bp},clip]{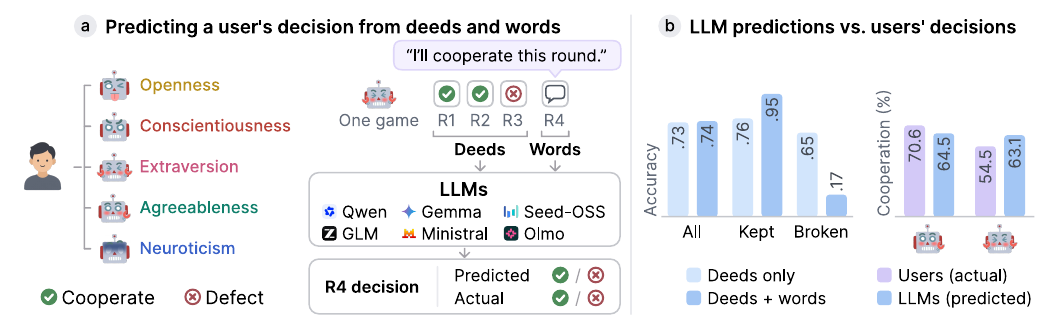}
\caption{LLM predictions varied with users' stated decisions but barely varied with the agent users faced, although users' behavior did. (a) Each user played a six-round game with each of five agents, each prompted to express a dominant Big Five trait, and could talk with the agent before deciding to cooperate or defect in every round. Eight open-weight LLMs predicted each decision from earlier decisions (deeds) and the round's dialogue (words). (b) Left: adding dialogue barely changed accuracy on rounds with a stated decision, but raised it on kept statements and lowered it on broken ones. Right: users cooperated more with the Agreeableness than with the Extraversion agent, whereas predictions from all decisions and dialogue barely differed.}
\Description{Two panels. (a) Left: a user linked to five agent personas, Openness, Conscientiousness, Extraversion, Agreeableness and Neuroticism. Right: one game with an agent; rounds 1 to 3 show the user's decisions (cooperate, cooperate, defect), labeled deeds, and round 4 shows the user saying I'll cooperate this round, labeled words. Both feed six LLM families (Qwen, Gemma, Seed-OSS, GLM, Ministral, Olmo), whose predicted round-4 decision is compared with the actual decision. (b) Left: accuracy with deeds only and with deeds and words: all rounds .73 and .74, kept .76 and .95, broken .65 and .17. Right: cooperation rate with the Agreeableness and Extraversion agents: users 70.6 and 54.5 percent, LLM predictions 64.5 and 63.1 percent.}
\label{fig:teaser}
\end{teaserfigure}

\maketitle

\section{Introduction}

Web systems predict what users will do next. Recommendation and ranking have long drawn these predictions from past behavior, such as clicks and purchases \citep{richardson2007predicting,hu2008collaborative,he2017neural}. Conversational interaction adds a more direct source, because users often say what they will do before they act. As LLM-based agents enter Web platforms, users talk with sales and service agents that represent another party and then decide whether to accept an offer or keep an agreement \citep{rothschild2026agentic,yang2025agentic}. When a user tells such an agent ``I will accept this offer,'' the dialogue appears to contain the answer to the prediction task.

User modeling has built the means to use such language. Conversational recommenders model users from the current dialogue \citep{jannach2021survey,he2023large,zhu2025collaborative}; LLMs predict individuals' choices from interviews \citep{park2024generative}, from earlier choices in psychology experiments \citep{binz2025foundation}, from play in persuasion games \citep{shapira2026labs} and after conversations \citep{pilli2026predicting}, and stand in for users when conversational agents are evaluated \citep{zhu2024how,zhu2025controllable,zhou2026mind,meshi2026convapparel}. Of the language available to a predictor, a user's own statement about the predicted action is the most direct evidence. A statement fixes the action the user names, but not how likely the user is to take it.

This gap matters most in mixed-motive interactions, where both sides gain from cooperation while their interests partly conflict: there a statement expresses a plan and also seeks to influence the other side. Promises raise cooperation in experimental games, yet about a quarter of them are broken \citep{charness2006promises,vanberg2008why}, and people broke more than half of their cooperation agreements with LLM agents \citep{wang2026llm}. A predictor therefore has to decide how far a statement should move its prediction, above all when the statement is inconsistent with the user's earlier decisions: ignoring it can miss real changes of plan, and relying on it heavily can overestimate how often it is kept. Overall accuracy does not reveal which error a predictor makes: a predictor that relies on statements gains accuracy on kept statements and loses it on broken ones, and the two can cancel. Whether dialogue improves prediction must therefore be judged against what users do later, separately for kept and broken statements.

A statement is also made to a particular counterpart, and the action is taken toward it. A counterpart's conduct and persona change people's decisions \citep{luo2019machines,chawla2023selfish,wang2026llm}, and how people respond to an agent's personality depends on their own: people were more attracted to a computer whose personality matched theirs \citep{moon1996how}, and human--AI teamwork depended on how the human's and the AI's traits were paired \citep{ju2026personality}. Users' decisions may therefore change with the agent's persona and with how that persona matches the user's own traits. A predictor that is accurate over all agents together can still be biased with particular agents or user--agent pairs.

Existing studies cannot settle either issue, as their data were collected for other questions. In choice-prediction studies, users receive messages from a bot \citep{shapira2026labs}, talk without stating the predicted action \citep{pilli2026predicting}, or play one-shot games with no record of earlier behavior \citep{gonzalezfernandez2026large}. User-simulator evaluations score generated conversation turns, so no later action can contradict what a simulated user says, and they compare responses to different assistants after pooling over users \citep{meshi2026convapparel}. No study links a user's statement to that user's earlier decisions, the counterpart, and the later action.

We therefore ask three research questions. 

\begin{itemize}[leftmargin=*,nosep]
    \item \textbf{RQ1:} How do LLM predictors combine users' stated decisions, what users say they will do in the current round, with the record of earlier decisions, and is their reliance on statements supported by how often statements are kept?
    \item \textbf{RQ2:} Do users' decisions and word-keeping vary with the agent persona, and do the predictions vary in the same way?
    \item \textbf{RQ3:} Do users' decisions vary with their own personality, measured as \emph{trait compatibility}, the user's score on the agent's defining trait, and do the predictions vary in the same way?
\end{itemize}

Answering them requires statements and actions recorded separately, repeated decisions that form a record against which to compare a statement, and counterparts that change while actions and payoffs stay the same. A repeated prisoner's-dilemma-style game with communication before each decision meets these requirements. We use the Personality-1260 dataset \citep{zhang2025rethinking}, in which 42 users played six rounds with each of five GPT-4o agents, each prompted to express a dominant Big Five trait, talking with the agent before every decision (Figure~\ref{fig:teaser}a). Eight open-weight LLMs predicted each decision from 19 combinations of the decision record (both players' earlier decisions, the \emph{deeds}) and the dialogue (what users said, the \emph{words}). We evaluate predictions separately for kept and broken statements and for those inconsistent with users' earlier decisions, measure the effect of one inserted statement, and compare them with users' decisions across agents and trait compatibility.

Predictions from the decision record separated cooperation from defection well (AUC = .80), and adding dialogue brought no detectable gain. Adding the round's dialogue left accuracy on rounds with a stated decision almost unchanged (.73 to .74), because it raised accuracy where users kept their statement (.76 to .95) and lowered it where they broke it (.65 to .17; Figure~\ref{fig:teaser}b, left). When users stated a decision they had not yet chosen in that game, they kept it at a rate of .40, while the predicted probability of keeping it rose from .13 without the dialogue to .69 with it: the predictions moved beyond what the statement warranted. Users cooperated in 70.6\% of rounds with the Agreeableness agent and 54.5\% with the Extraversion agent and kept more of their statements with the former; predicted cooperation was 64.5\% and 63.1\% (Figure~\ref{fig:teaser}b, right). Users also cooperated more as trait compatibility rose, whereas the predictions barely changed and fell increasingly below users' cooperation. Pooled over agents, the prediction bias was near zero.

This paper makes four contributions:
\begin{enumerate}[leftmargin=*,nosep]
    \item An evaluation design that ties a predictor's use of dialogue to separately recorded actions, split by whether statements were kept, whether they were inconsistent with users' earlier decisions, and the agent the user faced.
    \item Evidence that eight LLM predictors relied on stated decisions beyond how often they were kept.
    \item Evidence that the predictions missed how users' behavior varied with the agent persona and trait compatibility, and that pooled metrics concealed both errors.
    \item Design implications for dialogue-based user modeling: how much weight a user's statement receives should depend on the user's earlier decisions and on the agent, and predictions should be checked against what users later do with each agent.
\end{enumerate}

\section{Related Work}

\subsection{Prediction from Dialogue and History}

Studies of decision prediction have used behavioral history, conversational information \citep{tan2016winning}, or both. \citet{nay2016predicting} model cooperation in repeated games using the game structure and both players' previous actions. In language-based persuasion games, \citet{shapira2026labs} find that LLM-generated data support better human-choice prediction when they capture history-dependent behavior beyond the sentiment of the messages users receive. Dialogue can also provide context without revealing the user's intended choice: \citet{pilli2026predicting} predict decisions after cognitively demanding conversations, but replacing users' replies with random text changes accuracy little. In these settings, dialogue provides context for a choice without containing the user's statement of that choice. \citet{gonzalezfernandez2026large} examine such statements more directly by predicting strategic choices from pre-play communication. In their three games, GPT-5 achieved higher mean accuracy than human forecasters, but was much less accurate for defection than cooperation. Their one-shot settings, however, provide no repeated decision history against which a current statement can be assessed. Repeated interactions allow a further question: does a predictor interpret a current statement in light of the speaker's earlier choices and record of keeping statements?

\subsection{Stated Decisions as Behavioral Evidence}

A stated decision can be informative without guaranteeing that the stated action will occur. Promises can encourage trustworthy behavior \citep{charness2006promises}, and their wording matters: conditional or implicit statements in a high-stakes game show were associated with less cooperation than explicit, unconditional ones \citep{turmunkh2019malleable}. Yet even expressed agreement can precede a different action. In online donation dialogues, some participants who agreed to donate subsequently gave less or nothing \citep{wang2019persuasion}. We use \emph{stated decisions} to include expressions of intended action beyond explicit promises, without assuming that they will be kept. Recognizing that statements are informative does not ensure that an observer uses them appropriately. \citet{chen2017promises} found that human observers distinguished promises from empty talk, but overestimated promise keeping and could not reliably distinguish kept from broken promises. This separates two questions: whether statements predict behavior, and whether a forecaster's confidence in them matches observed behavior.

\subsection{Counterparts and User-Agent Pairing}

The behavior against which these forecasts are assessed also depends on the counterpart. \citet{demelo2011impact} found that emotional displays altered cooperation with agents using the same decision strategy. In conversational negotiation, agents' willingness to compromise affects whether people reach agreements \citep{chawla2023selfish}; with LLM counterparts, both cooperation and agreement keeping vary across personas \citep{wang2026llm}. These findings make the agent's conduct part of the context for predicting a user's next action.

Responses to that conduct can further depend on the user. Personality similarity increased attraction to computers in \citet{moon1996how}, whereas \citet{isbister2000consistency} found a preference for complementary interactive characters. More recently, \citet{ju2026personality} found that human and AI personality pairings affected teamwork and performance. We examine how the user's score on the agent's defining trait relates to decisions and prediction bias, without assuming that higher compatibility implies greater cooperation.

For prediction, the question is whether forecasts reproduce such contextual differences. ConvApparel tests whether simulated conversational responses reproduce real users' reactions to assistants of differing quality, finding that data-driven simulators adapt more faithfully than a prompted one \citep{meshi2026convapparel}. We bring this evaluation question to separately recorded actions after dialogue, comparing forecasts with behavior across agent personas and user-agent pairings.

\section{Experiment Design}

We study offline prediction on recorded games: for each round, a predictor receives material recorded before the user's decision and outputs the probability that the user cooperates. The agent played with the user; the predictor is a separate model, run afterwards, that only forecasts the user's recorded decision.

\begin{figure*}[t]
\centering
\includegraphics[width=\textwidth,trim={0 0.4bp 0 0.4bp},clip]{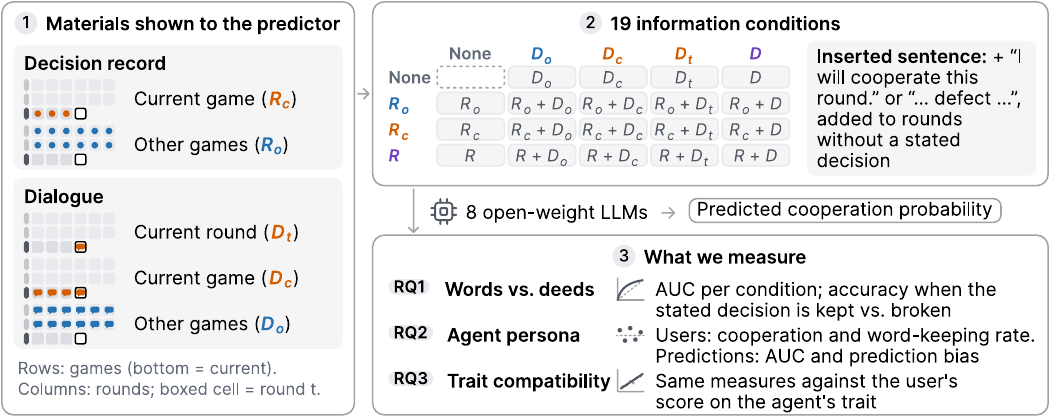}
\caption{Experiment design. \ding{192}~The five materials given to the predictor, defined by game and round. \ding{193}~The 19 combinations of materials shown to eight LLM predictors, and the sentence inserted into rounds without a stated decision. \ding{194}~What each research question measures in users' behavior and in the predictions.}
\Description{Three panels. Left: small grids of games by rounds showing the five materials given to the predictor. Middle: a four by five matrix of the 19 information conditions and the inserted sentence. Right: the measures used for each research question.}
\label{fig:design}
\end{figure*}

\subsection{Users, Agents and Game}

We use the Personality-1260 dataset \citep{zhang2025rethinking}. Forty-two participants from a Chinese university (21 female; mean age 22.1) completed the Chinese BFI-44 \citep{john1999big} and played one game with each of five agents in random order. In each of six rounds per game, user and agent may talk freely; then both decide whether to cooperate or defect, the agent's decision being revealed only after the user's. Mutual cooperation pays 2 points each, defecting against a cooperator 3 (the cooperator 0), and mutual defection 0 each. Talking was optional, and the dialogue was empty in 276 of the 1,260 rounds. 

Each agent is GPT-4o prompted to be extremely high in one Big Five trait, to maximize its score, and to show the trait in dialogue and decisions without naming it. GPT-4o expressed personas set to the high end of a trait more faithfully than other models \citep{shin2025spotting}, and blind experts rated each agent 4.53--4.80 of 5 on its trait \citep{zhang2025rethinking}. We call the agents by their \emph{agent persona}: Openness (O), Conscientiousness (C), Extraversion (E), Agreeableness (A), Neuroticism (N). Users cooperated in 64.5\% of rounds and agents in 77.7\% (Appendix~\ref{app:desc}).

\subsection{Stated Decisions}

A \emph{stated decision} is what the user said in the round's dialogue about their own decision in that round. Two researchers independently labeled each dialogue, shown without the decision, as stating cooperation, stating defection, or \emph{no stated decision}. Inter-rater agreement was high ($\kappa = .97$), and disagreements were resolved through discussion (Appendix~\ref{app:labels}). Statements were cheap: in games 2--5, 96\% of stated decisions announced cooperation, 27.6\% of stated cooperation was followed by defection, and 47.6\% of all defections came right after a stated cooperation.

\begin{figure*}[t]
\centering
\includegraphics[width=\textwidth,trim={0 12.6bp 0 4.1bp},clip]{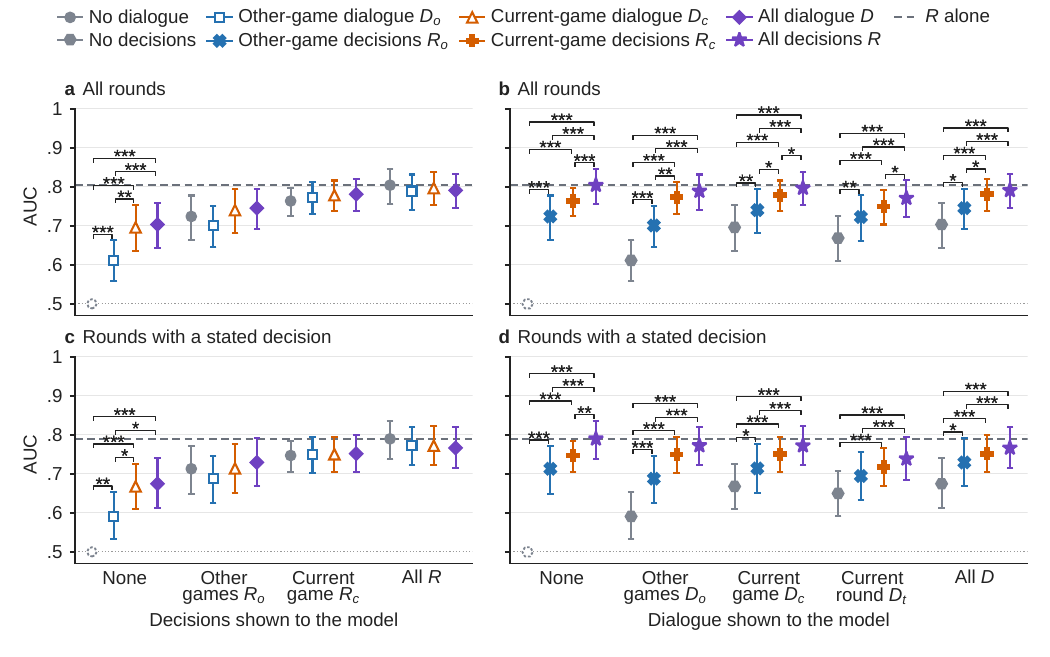}
\caption{The decision record carried the ranking information, and dialogue brought no detectable gain once the record was shown. AUC under each combination of decisions and dialogue shown to the model (games 2--5), grouped by decisions (a,~c) or by dialogue (b,~d). Dashed line: AUC of the decision record alone ($R$); dotted line: chance. Bars: 95\% user-bootstrap intervals; brackets: Holm-corrected tests. $^{*}p<.05$, $^{**}p<.01$, $^{***}p<.001$.}
\Description{Four panels of AUC with 95 percent intervals for the 19 information conditions. Conditions with a decision record reach AUC near .80; adding dialogue does not raise it further.}
\label{fig:rq1}
\end{figure*}

\subsection{Quantities}

Let $y_{uat} \in \{0,1\}$ be the \emph{decision} of user $u$ in round $t$ of the game with agent $a$ (1 = cooperate) and $s_{uat}$ the stated decision.

\paragraph{Users' behavior.} The \emph{cooperation rate} $\mathrm{CR}_{ua} = \frac{1}{6}\sum_t y_{uat}$ measures how cooperative a user was with one agent. The \emph{word-keeping rate} measures how reliable their words were with that agent,
\[
\mathrm{WK}_{ua} = \frac{1}{|S_{ua}|}\sum_{t \in S_{ua}} \mathbf{1}[y_{uat} = s_{uat}],
\]
where $S_{ua}$ holds the rounds with a stated decision; a matched round is \emph{kept}, otherwise \emph{broken}. A \emph{relative} rate subtracts the user's own mean over their games (e.g., $\mathrm{CR}_{ua} - \frac{1}{5}\sum_{a'} \mathrm{CR}_{ua'}$), isolating what one agent changed in that user's behavior.

\paragraph{Trait compatibility.} $c_{ua}$ is the user's BFI-44 score on the trait agent $a$ expresses (e.g., the user's Agreeableness with the Agreeableness agent), standardized within each agent; it measures how far the agent's defining trait is also the user's own. We use this one trait rather than similarity across all five because traits are expressed in response to situational cues relevant to them \citep{tett2003personality}, and each agent was prompted to express one dominant trait.

\paragraph{Predictions.} For model $m$ and answer-letter order $k \in \{1,2\}$ (cooperate as A or as B), the score is the next-token log-odds $\ell^{(k)}_m = \log p(\text{cooperate letter}) - \log p(\text{defect letter})$. The average $\ell_m = (\ell^{(1)}_m + \ell^{(2)}_m)/2$ cancels a preference for one letter \citep{zheng2024large}. The \emph{predicted cooperation probability} is $p_m = \sigma(\ell_m)$; the \emph{predicted word-keeping probability} is $p_m$ after a stated cooperation and $1 - p_m$ after a stated defection. Each quantity is computed per model and averaged over the eight models (\emph{8-model mean}).

\paragraph{Evaluation.} \emph{AUC}, the probability that a random cooperation round scores above a random defection round, measures how well predictions rank decisions and ignores their level. \emph{Accuracy} is the share of rounds with $\mathbf{1}[\ell_m > 0] = y$. The \emph{prediction bias} of a game, $b_{ua} = 100\,(\overline{p}_{ua} - \overline{y}_{ua})$, is the mean predicted probability minus the \emph{observed rate} over the same rounds, in points (all rounds for cooperation; rounds with a stated decision for word-keeping).

\subsection{Predictors and Information Conditions}

Eight open-weight instruction-tuned LLMs from six families served as predictors: Qwen3.5-9B, Qwen3.8-27B, Gemma-4-12B, Gemma-4-31B, Seed-OSS-36B, GLM-4-32B, Ministral-3-14B and Olmo-3.1-32B (Appendix~\ref{app:models}). Each received, in Chinese like the dialogues, the rules and payoffs, the round number, a combination of materials, and a question with the two options as answer letters (Appendix~\ref{app:prompt}), with reasoning disabled. The materials come from the user's \emph{decision record} ($R$) and the dialogue ($D$), with subscripts for where they come from (Figure~\ref{fig:design}): $R_c$ holds both players' decisions in earlier rounds of the current game and $R_o$ those in the user's earlier games with other agents; $D_t$ is the dialogue of the current round $t$ before the decision, $D_c$ the dialogue of the current game up to that point (earlier rounds and round $t$), and $D_o$ the dialogue of the user's other games. Without a subscript, $R$ and $D$ include everything, and a condition is written as the sum of what it shows (e.g., $R_c + D_t$). Decisions shown had four levels (none, $R_o$, $R_c$, $R$) and dialogue five (none, $D_o$, $D_c$, $D_t$, $D$): 19 conditions, the empty cell not run. Prediction analyses use games 2--5 (1,008 rounds), as other-game materials need an earlier game, and users' behavior is compared with the predictions on the same rounds. Analyses of users' behavior alone across agents use all five games (1,260 rounds), as relative rates are taken within each user (Appendix~\ref{app:subsets} lists the rounds behind each finding).

\paragraph{Inserted sentences.} In condition $R_c + D_t$ we appended ``User: I will choose to cooperate this round.'' (in Chinese) to the current-round dialogue of rounds 2--6 without a stated cooperation, and ``\ldots defect \ldots'' to those with no stated decision; separately, we swapped the user's previous decision in the prompt. The \emph{weight ratio} divides the log-odds change caused by the sentence by that caused by the swap; its behavioral counterpart is the ratio of the corresponding coefficients in a logistic regression of users' decisions (Appendix~\ref{app:stats}).

\begin{figure*}[t]
\centering
\includegraphics[width=\textwidth,trim={0 6.8bp 0 4.1bp},clip]{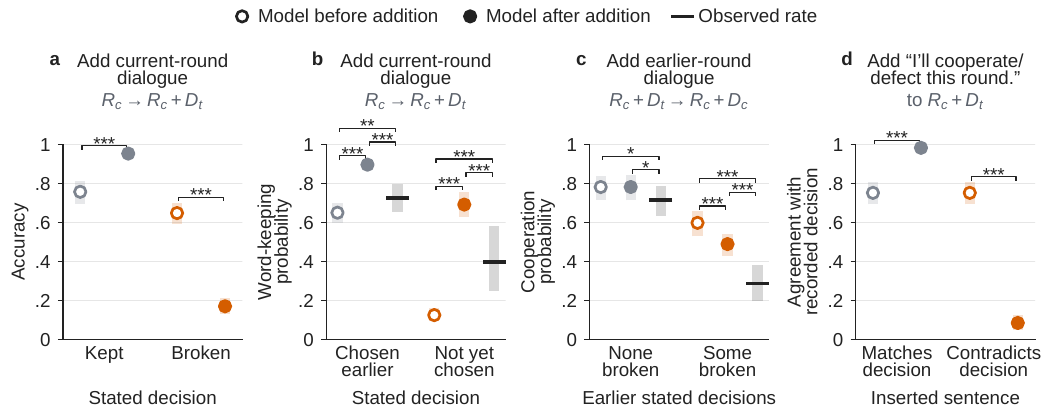}
\caption{Adding what users said moved the predictions toward their statements, also where users then broke them (a) and beyond the observed rate of keeping them (b), while earlier broken statements were only partly used (c); one inserted sentence set the prediction (d). Accuracy, agreement with the recorded decision (d) or predicted probability before (open) and after (filled) adding text to the prompt (games 2--5; group sizes in Appendix~\ref{app:subsets}); black bars: observed rates. Light bars: 95\% user-bootstrap intervals; brackets: Holm-corrected tests. $^{*}p<.05$, $^{**}p<.01$, $^{***}p<.001$.}
\Description{Four dot plots of model accuracy or predicted probability before and after adding text to the prompt. Adding the current-round dialogue raises accuracy where the stated decision was kept and lowers it sharply where it was broken.}
\label{fig:rq1b}
\end{figure*}

\subsection{Statistical Analysis}

Rounds are nested in games and games in users, so uncertainty comes from a user bootstrap: 50,000 draws of 42 users with replacement, each bringing all their rounds. We report percentile intervals and two-sided bootstrap $p$-values, Holm-corrected within each family of tests displayed together (Appendix~\ref{app:stats}); a model difference counts as significant only if it has the same sign under each letter order. For RQ3 we estimate the least-squares slope of a game-level outcome on compatibility, pooling agents, without correction.

\section{Results}

\subsection{RQ1: Words and Deeds}

\finding{1}{Predictions from earlier decisions separated cooperation from defection better than predictions from dialogue, and adding dialogue to earlier decisions brought no detectable gain.}
Decisions alone ($R$) reached AUC = .80 [.76, .85], dialogue alone ($D$) .70 [.64, .76], and both ($R + D$) .79 [.75, .83] (Figure~\ref{fig:rq1}a,~b). Adding a decision record to any level of dialogue raised AUC in all 12 comparisons (+.04 to +.18, all $p_{\text{Holm}} < .05$), whereas within each non-empty level of decisions, the conditions with no dialogue, $D_o$, $D_c$ and $D$ did not differ from one another (0 of 18 comparisons). The pattern largely held in the 646 rounds with a stated decision (11 of 12 and 0 of 18; Figure~\ref{fig:rq1}c,~d). Even dialogue from other games ranked current decisions above chance ($D_o$: .61): what users say carries information about their decisions, which the decision record largely already provides. The stated decision itself ranks poorly: with 96\% of statements announcing cooperation, the statement separates cooperation from defection at AUC = .55.

\begin{figure*}[t]
\centering
\includegraphics[width=\textwidth,trim={0 2.9bp 0 0.9bp},clip]{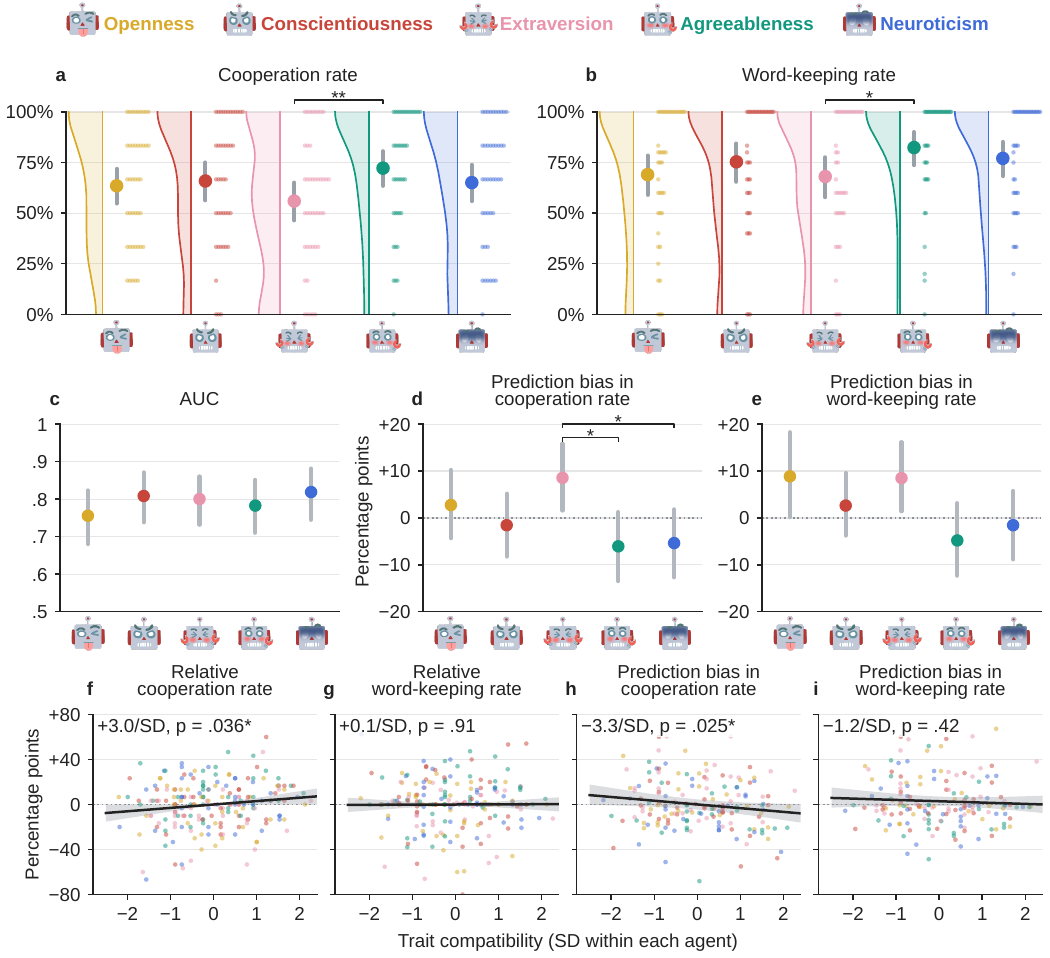}
\caption{Users' decisions shifted with the agent persona and with trait compatibility; the predictions barely varied with either. a,~b: users' rates by persona (all games; dots are games). c--e: AUC and prediction bias by persona (games 2--5). f--i: relative rates and prediction bias against trait compatibility (dots: games; lines: pooled slopes with 95\% user-bootstrap bands). Bars: 95\% user-bootstrap intervals; brackets: Holm-corrected tests. $^{*}p<.05$, $^{**}p<.01$.}
\Description{Nine panels. Top two rows by agent persona: users' cooperation and word-keeping rates differ between personas, AUC is similar across personas, and prediction bias is positive for the Extraversion agent and negative for the Agreeableness agent. Bottom row: scatter plots against trait compatibility with pooled slopes.}
\label{fig:rq23}
\end{figure*}

\finding{2}{Similar overall accuracy concealed opposite changes; adding the round's dialogue raised accuracy where users kept their statements and lowered it where they broke them.}
Adding the current-round dialogue to the current-game decisions changed accuracy on rounds with a stated decision from .73 to .74. It raised accuracy on the 470 rounds where the stated decision was kept from .76 to .95 (+.20 [.15, .25]) and lowered it on the 176 rounds where it was broken from .65 to .17 ($-$.48 [$-$.52, $-$.44]; both $p_{\text{Holm}} < .001$; Figure~\ref{fig:rq1b}a). Because kept rounds outnumber broken ones, the gains offset the losses: the dialogue moved the errors onto the rounds in which users broke their word, which held nearly half of all defections. Where the errors fall does not yet show whether the predictions gave the statement too much weight; the rounds in which a statement is inconsistent with the user's earlier decisions show this directly.

\finding{3}{When users' stated decisions were inconsistent with their earlier decisions, the predictions overestimated how often they kept them. Evidence that users had broken earlier statements was only partly used.}
When users stated a decision they had already chosen earlier in the game, the round's dialogue raised the predicted word-keeping probability from .65 to .90, against an observed .73 (Figure~\ref{fig:rq1b}b). When they stated one they had not yet chosen (53 rounds, 20 users), the prediction from the current-game decisions alone was .13 and rose to .69 with the dialogue, while users kept their word in .40 [.25, .58] of these rounds: the prediction moved from .27 below the observed rate to .30 above it (both $p_{\text{Holm}} < .001$). Among users who had broken a stated decision earlier in the game, adding the earlier-round dialogue, which contains the broken statement, lowered predicted cooperation from .60 to .49, against an observed .29 (Figure~\ref{fig:rq1b}c). The statement was informative, since users kept it far more often than predicted without it, and the predictions gave it more weight than its keeping rate supported. Evidence that a user's words had failed moved the predictions by only eleven percentage points.

\finding{4}{One sentence stating a decision moved the predictions more than statements went with actions in users' behavior.}
In rounds 2--6 without a stated decision, one inserted sentence moved agreement with the recorded decision from .75 to .98 when it matched that decision and to .08 when it did not (Figure~\ref{fig:rq1b}d). Relative to swapping the user's previous decision, the inserted cooperation sentence moved the log-odds .63 to 2.60 times as much across models (mean 1.34); in users' behavior, the \mbox{coefficient} of a stated cooperation was .35 [.11, .68] times that of the previous decision. The difference was positive for all eight models (mean +.99 [.65, 1.38], $p_{\text{Holm}} < .001$; significant for six). The first ratio measures a response to an input change and the second an association in behavior; their comparison shows that the predictions moved with statements more than users' actions went with them.

\subsection{RQ2: Agent Persona}

\finding{5}{Users cooperated and kept their word most with the Agreeableness agent and least with the Extraversion agent. How often they stated a decision did not differ detectably.}
Users cooperated in 72.2\% [63.5, 80.6] of rounds with the Agreeableness agent and 56.0\% [46.4, 65.1] with the Extraversion agent (+16.3 [6.7, 25.8], $p_{\text{Holm}} = .008$; Figure~\ref{fig:rq23}a) and on average kept 82.4\% versus 68.1\% of their stated decisions (+14.3 [4.7, 24.5], $p_{\text{Holm}} = .022$; Figure~\ref{fig:rq23}b). No other pair differed after correction, nor did any pair in how often users stated a decision. The difference is not explained by the agents' own cooperation alone: the Neuroticism agent cooperated least (67.5\%), yet users cooperated with it descriptively more than with the E agent (65.1\% vs.\ 56.0\%).

\finding{6}{The predictions separated cooperation from defection equally well with every agent. They varied less across agents than users did, and their bias differed between agents.}
In games 2--5, observed cooperation differed by 16.1 points between the Agreeableness and Extraversion agents (70.6\% vs.\ 54.5\%), whereas predictions with all information ($R + D$) differed by 1.4 points (64.5\% vs.\ 63.1\%); across all personas, predicted cooperation spanned 6.0 points. AUC ranged from .76 (O) to .82 (N), and no pair differed (Figure~\ref{fig:rq23}c). The prediction bias in cooperation rate was +8.6 [1.6, 15.8] points with the Extraversion agent and $-$6.0 [$-$13.5, 1.3] with the Agreeableness agent (Figure~\ref{fig:rq23}d), differing between them (14.6 points, $p_{\text{Holm}} = .013$) and between E and N (13.9, $p_{\text{Holm}} = .018$). No single persona's bias differed from zero after correction. The prediction bias in word-keeping rate showed the same order (E +8.5, A $-$4.8; $p_{\text{Holm}} = .057$; Figure~\ref{fig:rq23}e). Pooled over games 2--5, the bias in cooperation rate was $-$0.1 [$-$4.6, 4.4] points.

AUC measures how well the predictions separate cooperation from defection and ignores their level; the decision records allowed this separation with every agent. The agent changed the level, and the predictions stayed at one level, although the agent's decisions and dialogue were in the prompt. The near-zero pooled mean bias concealed differences of up to 14.6 points between agents.

\subsection{RQ3: User Personality and Compatibility}

\finding{7}{The higher users themselves scored on the trait an agent expressed, the more they cooperated with that agent. We detected no such dependence for keeping their word.}
The relative cooperation rate rose by 3.0 points per standard deviation of compatibility [0.2, 6.0] ($p = .036^{*}$; Figure~\ref{fig:rq23}f); the relative word-keeping rate did not change (+0.1, $p = .91$; Figure~\ref{fig:rq23}g).

\finding{8}{The predictions did not vary with compatibility and fell increasingly short of users' cooperation as the agent's trait matched the user's. Telling the LLMs users' personality scores closed only part of this gap.}
The prediction bias in cooperation rate fell by 3.3 points per standard deviation [$-$6.4, $-$0.4] ($p = .025^{*}$, same sign under both letter orders; Figure~\ref{fig:rq23}h): in games 2--5, the observed cooperation rate rose by 4.1 points per standard deviation and the predicted one by 0.8, and the bias slope was negative within each of the five agents ($-$2.2 to $-$5.4). The prediction bias in word-keeping rate showed no slope ($-$1.2, $p = .42$; Figure~\ref{fig:rq23}i). To test whether the bias arose because the predictors lacked this information, we added the user's BFI scores, or the compatibility score itself, to the prompt with all decisions and dialogue (Appendix~\ref{app:traitinfo}). The slope was then $-$2.5 points per standard deviation in both cases ([$-$5.3, 0.3], $p = .082$; [$-$5.6, 0.4], $p = .090$), less negative by 0.9 [0.2, 1.5] ($p = .005$) and 0.8 [0.4, 1.2] ($p < .001$; same sign under both letter orders), and the compatibility score changed it no more than the BFI scores did ($-$0.1 [$-$0.6, 0.4], $p = .83$): most of the bias remained with the feature in the prompt.

\section{Discussion}

\subsection{Grounding Statements in Behavior}
\label{sec:disc-words}

A user can state an action unambiguously while leaving substantial uncertainty about whether it will occur. When users named an action they had not previously chosen, adding the current-round dialogue moved mean predicted fulfillment from below the observed rate to above it (Finding~3). Returning to the history-only condition would leave the original underestimation unresolved. The central issue is therefore how far dialogue revises a forecast and whether that revision is supported by subsequent behavior. The insertion experiment establishes sensitivity to a stated decision; naturally occurring statement-action pairs provide the behavioral reference needed to assess that sensitivity. Human observers have also overestimated promise keeping \citep{chen2017promises}. Our repeated setting exposes a further difficulty: even when earlier unfulfilled statements became available, predictions remained above observed cooperation (Finding~3). An action log alone cannot reveal whether an earlier statement was kept, and the earlier dialogue contains the statements without their outcomes. We therefore added to the prompt a record of what the user had said in each earlier round and what they chose (Appendix~\ref{app:keeprecord}). For statements inconsistent with the user's earlier decisions, the predicted word-keeping probability fell from .69 to .40, the observed rate; after an earlier broken statement, predicted cooperation fell from .60 to .40 (observed .29). Adding the earlier-round dialogue instead lowered them only to .60 and .49 (both differences from the record $p_{\text{Holm}} < .001$). Linking each earlier statement to its outcome corrected most of the overestimation in Finding~3. A sales or service agent that decides whether to grant a concession on a user's promise could therefore keep, for each user, a record of what they said and then did, and weigh a new promise against it.

\subsection{Modeling Users with Their Counterparts}

The behavioral evidence itself depends on the interaction. Users cooperated and fulfilled statements at different rates across agent personas (Finding~5), and cooperation was associated with trait compatibility (Finding~7). A user representation that absorbs these choices into a general cooperation tendency risks losing information about the counterpart involved. Compatibility also showed no detected association with statement fulfillment: willingness to cooperate and the reliability of a stated decision remain distinct outcomes. These results support evaluating a user's action in relation to the counterpart and interaction history, without treating a single game as evidence of a stable personal disposition.

Making that context available is only one part of the problem. The main persona comparison already supplied the counterpart's earlier actions and dialogue. Adding the user's BFI scores or the compatibility score reduced the estimated compatibility-related bias slope by roughly one quarter (Finding~8). The remaining point estimates were negative, although their intervals included zero. The added information therefore helped, while the extent of the remaining error is uncertain. Richer user descriptions require validation of how they change predicted behavior; their presence in a prompt provides no assurance that its forecasts reproduce the relevant behavioral differences. The contrast with the word-keeping record (Section~\ref{sec:disc-words}) suggests that a predictor uses added information best when it states the relation to be predicted: the record linked each statement to its outcome, whereas the trait scores left the predictor to infer how a trait shapes behavior toward a particular agent. A platform deploying agents with different personas should therefore calibrate predictions per persona instead of reusing one user profile: a user who keeps agreements with one agent may not with another.

\subsection{Evaluating Errors That Averages Conceal}

Two different cancellations made overall performance appear reassuring. Gains on fulfilled statements offset losses on unfulfilled statements, while positive and negative mean biases across personas nearly canceled (Findings~2 and~6). A different mixture of fulfillment outcomes or counterparts could therefore change aggregate performance even with unchanged within-group performance. Likewise, AUC measures discrimination between cooperation and defection, leaving the agreement between predicted probabilities and observed rates to be checked separately. Together, these results show how a model can perform well on aggregate measures while missing relationships that matter within particular interactions.

Evaluation should consequently connect forecast changes to independently recorded actions under identifiable interaction conditions. Splitting by subsequent fulfillment locates errors retrospectively. Testing whether those errors are recognizable before a decision requires groups defined from available evidence, such as a statement inconsistent with the user's earlier decisions or an earlier unfulfilled statement. Comparisons across counterparts and levels of compatibility then test whether predictions reproduce contextual variation in behavior. These checks can also guide evaluation of action-generating user simulators, extending the concern with responsiveness to different assistants in ConvApparel \citep{meshi2026convapparel}. Matching a pooled cooperation rate alone would provide little evidence that a simulator reproduces users' responses to a particular agent. Before an LLM predictor or user simulator is deployed, its accuracy should therefore be reported separately for kept and broken statements and for each agent, as a pooled score can look adequate while it fails on the users and agents that matter most.

\subsection{Limitations and Future Work}

We studied 42 Chinese university students in short, low-stakes games and eight zero-shot predictors using Chinese prompts with reasoning disabled. Generalization to other populations, longer relationships, and Web tasks needs validation. All counterparts used GPT-4o; persona prompts changed language and decisions, confounding personality expression with behavioral reciprocity. Compatibility uses one trait; analyses are exploratory and uncorrected. Some subgroups were small, including 53 rounds from 20 users. Prompt edits measure model sensitivity, while statement-action associations do not identify causal effects. Future work should independently vary agent language and decision policy, and test history-aware or counterpart-conditioned predictors on held-out users and interactions across prompts, reasoning settings, and trained models.

\section{Conclusion}

In this paper, we present an empirical evaluation of how LLMs combine dialogue and behavioral history to predict users' decisions. Using recorded mixed-motive games with 42 users and five agent personas, we examined eight open-weight predictors through information ablations, statement insertions, and analyses of trait compatibility. The predictors overestimated the fulfillment of stated decisions and understated behavioral variation associated with agent persona and trait compatibility. Nearly unchanged accuracy on stated-decision rounds and near-zero mean bias pooled across agents obscured these mismatches. Supplying explicit personality information modestly reduced the estimated compatibility-related bias slope. Together, these findings show why dialogue-based user models need evaluation that connects their use of language and interaction context to subsequent actions. We hope this work supports user models that connect what people say to how they have acted and to the counterpart they face, while preserving uncertainty about choices they have yet to make.

\bibliographystyle{ACM-Reference-Format}
\bibliography{references}

\appendix

\section{Predictor Models and Inference}
\label{app:models}

\begin{table}[h]
\centering
\resizebox{\columnwidth}{!}{%
\begin{tabular}{@{}ll@{}}
\toprule
Model & Hugging Face identifier \\
\midrule
Qwen3.5-9B & \texttt{Qwen/Qwen3.5-9B} \\
Qwen3.8-27B & \texttt{Qwen/Qwen3.8-27B} \\
Gemma-4-12B & \texttt{google/gemma-4-12B-it} \\
Gemma-4-31B & \texttt{google/gemma-4-31B-it} \\
Seed-OSS-36B & \texttt{ByteDance-Seed/Seed-OSS-36B-Instruct} \\
GLM-4-32B & \texttt{zai-org/GLM-4-32B-0414} \\
Ministral-3-14B & \texttt{mistralai/Ministral-3-14B-Instruct-2512-BF16} \\
Olmo-3.1-32B & \texttt{allenai/Olmo-3.1-32B-Instruct} \\
\bottomrule
\end{tabular}}
\caption{Predictor models.}
\label{tab:models}
\end{table}

All models ran in vLLM with bf16 weights, one prompt at a time (no batching, no prefix caching), with each model's chat template and a single user message; reasoning was disabled where a template supports it. We generated one token greedily and read the log-probabilities of the two answer letters from the top-20 candidates, with an exact prompt-log-probability pass as a fallback. For GLM-4-32B, whose template always begins its reply with a newline, the reply was prefilled with the Chinese word for ``Answer:''. Inference ran on NVIDIA L20 and RTX A6000 GPUs.

\section{Prompt Template}
\label{app:prompt}

Prompts were in Chinese; below is an English translation of the fixed parts. Blocks absent from a condition are left out.

\begin{promptbox}{Prompt template}
Below is the record of a prisoner's dilemma game. A human player and an AI agent play 6 rounds. In each round the two first talk, then each secretly chooses "cooperate" or "defect". Scoring: both cooperate, 2 points each; one defects and one cooperates, the defector gets 3 and the cooperator 0; both defect, 0 each. In each round the agent decides first; the player cannot see the agent's choice for this round. In the dialogue, "User" is the player and "Agent" is the AI agent.

\cond{$R_o$ and/or $D_o$} Before this game, this player has played one game with each of several other AI agents ... Below are both players' decisions \cond{and the dialogue} in every round of those games. This is game \{order\} for this player.

This is round \{t\} of 6 of the current game.

\cond{$R_c$ and/or $D_c$} Earlier rounds of this game: Round \{k\} dialogue: ... Round \{k\} result: the player chose \{...\}, the agent chose \{...\}.

\cond{$D_c$ or $D_t$} Round \{t\} (current round) dialogue: ...

What will the player choose in this round?\\
A. Cooperate\quad B. Defect\\
Answer with one letter only.
\end{promptbox}

\begin{table*}[t]
\centering
\setlength{\tabcolsep}{12pt}
\begin{tabular}{@{}lrrl@{}}
\toprule
Subset & Rounds & Users & Findings \\
\midrule
All rounds, games 1--5 & 1,260 & 42 & 5, 7 \\
All rounds, games 2--5 & 1,008 & 42 & 1, 6, 8 \\
With a stated decision (Fig.~\ref{fig:rq1b}a) & 646 & 41 & 1, 2 \\
\quad kept / broken & 470 / 176 & 41 / 31 & \\
Rounds 2--6 with a stated decision (Fig.~\ref{fig:rq1b}b) & 532 & 41 & 3 \\
\quad chosen earlier / not yet chosen & 479 / 53 & 41 / 20 & \\
Rounds 2--6 (Fig.~\ref{fig:rq1b}c; regression) & 840 & 42 & 3, 4 \\
\quad no / some statement broken earlier & 631 / 209 & 42 / 28 & \\
Rounds 2--6, no stated decision (Fig.~\ref{fig:rq1b}d) & 308 & 40 & 4 \\
Rounds 2--6, no stated cooperation (weight ratio) & 327 & 40 & 4 \\
\bottomrule
\end{tabular}
\caption{Rounds and users in each analysis subset.}
\label{tab:subsets}
\end{table*}

\section{Personality Information in the Prompt}
\label{app:traitinfo}

Two further conditions each added one block to the prompt with all decisions and dialogue ($R + D$), placed after the rules; both were run with the eight models on the 1,008 rounds of games 2--5 in both letter orders. The first block gave the user's BFI-44 scores, the second the compatibility score without naming the trait (translated; $\{pct\}$ is the share of the other 41 users scoring lower on that trait):

\begin{promptbox}{BFI-44 scores}
This player completed a Big Five personality questionnaire (self-report). Each dimension is scored from 1 to 5; higher scores indicate a stronger trait. The means of the 42 players in this study are in parentheses. Openness: \{O\} (mean 3.47); Conscientiousness: \{C\} (mean 3.13); Extraversion: \{E\} (mean 3.10); Agreeableness: \{A\} (mean 3.76); Neuroticism: \{N\} (mean 2.95).
\end{promptbox}

\begin{promptbox}{Compatibility score}
The AI agent in this game was set to be extremely high on one Big Five dimension. On the same dimension, this player's questionnaire score is \{score\} (self-report, 1--5), higher than \{pct\}\% of the players in this study.
\end{promptbox}

Tests follow Finding~8, which reports the results.

\section{Word-Keeping Record in the Prompt}
\label{app:keeprecord}

One further condition added to $R_c + D_t$ a block listing, for each earlier round in which the user stated a decision, what the user said and what they then chose. The block came after the earlier-round results and before the current-round dialogue; round 1 had none. It was run with the eight models on the 1,008 rounds of games 2--5 in both letter orders. Translation (\{said\} and \{did\} are ``cooperate'' or ``defect''; rounds without a stated decision are omitted):

\begin{promptbox}{Word-keeping record}
In the earlier rounds of this game, the decisions this player stated in the dialogue and the decisions they actually made:\\
Round \{k\}: said they would \{said\}; actually chose \{did\}.\\
...\\
In total: stated a decision \{n\} times; the actual decision matched the statement \{j\} times.

\cond{no earlier statement} In the earlier rounds of this game, this player did not state a decision in the dialogue.
\end{promptbox}

The tests compare the record with the earlier-round dialogue ($R_c + D_c$) in the groups of Figure~\ref{fig:rq1b}b and~c, with the same user bootstrap, Holm correction and letter-order rule as that figure. For statements consistent with earlier decisions and users with no broken statement, the record changed predictions by less than .05.

\section{Stated-Decision Labels}
\label{app:labels}

Two researchers independently read each round's dialogue without seeing the user's decision and assigned a label based on the user's words, using the agent's words only to understand what the user was responding to. The five coding categories were: \emph{explicit cooperation} (the user says they will cooperate), \emph{implicit cooperation} (the user accepts the agent's proposal to cooperate or clearly commits without naming the decision), \emph{defection} (the user says or threatens that they will defect, including in jest), \emph{vague} (the user discusses the decision without committing, or only asks the agent), and \emph{none} (the user does not discuss their own decision this round, including talk about earlier rounds only). If the user changed their statement, the last statement counted. Across the 1,260 rounds, the final resolved labels comprised 594 explicit cooperation, 197 implicit cooperation, 30 defection, 116 vague, and 323 none. Inter-rater agreement was $\kappa = .81$ for the five-class coding scheme and $\kappa = .97$ after collapsing explicit and implicit cooperation into \emph{stated cooperation}, retaining defection as \emph{stated defection}, and merging vague and none into \emph{no stated decision}. All disagreements were resolved through discussion.

\section{Descriptive Statistics}
\label{app:desc}

Users cooperated in 64.5\% of the 1,260 rounds, falling from 82.4\% in round 1 to 42.4\% in round 6. Agents cooperated in 77.7\% of rounds: O 78.2\%, C 81.3\%, E 72.2\%, A 89.3\%, N 67.5\%. After removing decision-submission records, the dialogue was empty in 276 rounds; in the other 984 rounds users spoke 2.4 turns on average. Users' mean BFI-44 scores (1--5) were O 3.47, C 3.13, E 3.10, A 3.76, N 2.95. Because game order was random, the number of users who met each persona in games 2--5 ranged from 30 (O, A) to 39 (C). Overall, users kept 74.3\% of their stated decisions (74.0\% of stated cooperation and 83.3\% of stated defection).

\section{Analysis Subsets}
\label{app:subsets}

Table~\ref{tab:subsets} lists the rounds and users behind each analysis. Prediction analyses use games 2--5, as other-game materials need an earlier game, and whenever users' behavior is compared with the predictions, it comes from the same rounds. Analyses of users' behavior alone across agents (Findings~5 and~7) use all five games, as relative rates use each user's five games. Analyses comparing a round with earlier rounds of its game (Figure~\ref{fig:rq1b}b--d) exclude round 1.

\section{Test Families and Weights of the Inserted Sentence}
\label{app:stats}

\paragraph{Test families.} Holm correction was applied within: each group of a Figure~\ref{fig:rq1} panel (6 tests); each Figure~\ref{fig:rq1b} panel (3 or 7 tests, including the contrast between rows); the ten pairs of personas for each measure in Figure~\ref{fig:rq23}; the five tests of each persona's prediction bias against zero; and the three tests of the inserted sentence in each base condition (weight-ratio difference, change of the previous-decision effect, change of AUC).

\paragraph{Weight in the predictions.} For each model, $S$ is the mean change in $\ell_m$ when the cooperation sentence is appended (327 rounds), and $H$ the mean difference in $\ell_m$ between the prompt stating that the user cooperated in the previous round and the prompt stating that they defected, obtained by swapping that one line (same rounds, no inserted sentence). The model's weight ratio is $S/H$; the reported ratio is the mean over the eight models (per-model ratios .63--2.60, all above the behavioral ratio). The 327 rounds include 19 with a stated defection, which the inserted sentence contradicts; the behavioral coefficient contrasts stated cooperation with no statement.

\paragraph{Weight in behavior.} Over rounds 2--6 of games 2--5 (840 rounds), a logistic regression of the user's decision on stated cooperation, stated defection, the user's previous decision and the agent's previous decision gave coefficients of 0.67 for stated cooperation and 1.91 for the user's previous decision, a ratio of .35. Both ratios are recomputed within each of the 50,000 bootstrap draws.

\paragraph{Second wording.} A second inserted sentence (``I'll pick cooperate this round''), tested for cooperation only, had a similar effect.

\section{Ethical Considerations}

We reuse Personality-1260, whose collection received Institutional Review Board approval and written informed consent, reconfirmed after debriefing; participants received US\$10 \citep{zhang2025rethinking}. The dialogues contain no identifying information, and our predictors ran on local servers. An unfulfilled statement can reflect a changed decision and should not be used to label a person dishonest. Better prediction and persona selection also create risks of manipulative targeting. Applications should disclose agent roles and data use, preserve users' ability to revise their choices, and obtain authorization separately from any forecast of what a user might do.

\end{document}